\documentclass{bmvc2k}
\usepackage{booktabs}
\usepackage{amsfonts}
\usepackage{amsmath}
\usepackage{amssymb}
\usepackage{nicefrac}
\usepackage{microtype}
\usepackage{graphicx}
\usepackage[table]{xcolor}
\definecolor{ourshl}{RGB}{255,235,235}
\usepackage{graphicx}
\usepackage{multirow}
\usepackage{subcaption}
\DeclareMathOperator*{\argmax}{arg\,max}
   \usepackage{algorithm}
   \usepackage{algpseudocode}
\title{CrACK: Adversarial Attacks on Cross-Model Consistency\\in Collaborative Vision Foundation Models}

\addauthor{Feifei Liu}{20238331056@m.scnu.edu.cn}{1}
\addauthor{Jintao Cheng}{jchengau@connect.ust.hk}{2}
\addauthor{Chi Man VONG}{cmvong@um.edu.mo}{3}
\addauthor{Xiaoyu Tang$^{*}$}{tangxy@scnu.edu.cn}{1}

\addinstitution{
  the School of Data science and Engineering, South China Normal University\\
  $^{*}$Corresponding Author
}

\addinstitution{Hong Kong University of Science 
\\and Technology}

\addinstitution{University of Macau}
\runninghead{Liu et al.}{CrACK: Adversarial Attacks}

\begin{document}
\maketitle
\begin{abstract}
Training-free collaborative pipelines that integrate Vision Foundation
Models such as CLIP, SAM, and DINO achieve strong open-vocabulary dense
prediction and are increasingly deployed in safety-critical
applications. The security of these systems is commonly assumed to
follow from the robustness of their individual models. We challenge
this assumption. We identify a vulnerability shared by every
collaborative pipeline: each model consumes the intermediate output of
another without verifying semantic consistency, an unverified premise
that we term the semantic-spatial alignment dependency. Existing
adversarial attacks target a single model and overlook this premise,
leaving the inter-model interface entirely unguarded. We propose CrACK
(Cross-model Adversarial Consistency attack), an inference-time attack
that exploits this interface without modifying any input pixel, model
weight, or training data. CrACK operates in two stages: Adversarial
Affinity Contradiction Injection corrupts the cross-modal affinity
matrix by inverting SAM encoder features under the guidance of CLIP
patch-level semantics, and Semantic Interface Poisoning steers the
prediction through a max-distance label permutation derived from CLIP
text embeddings. Experiments on four collaborative pipelines across
eight benchmarks show that CrACK causes catastrophic degradation while
every individual model continues to produce its unchanged standalone
output, rendering per-model defenses structurally blind. The corruption
further cascades into large vision-language model reasoning, driving
models such as LLaVA to produce erroneous responses from visually
intact inputs. Our results show that the security of a collaborative AI
system cannot be reduced to the robustness of its components, and that
inter-model feature interfaces must be treated as first-class security
boundaries.
\end{abstract}

\section{Introduction}
\label{sec:intro}
 
Vision Foundation Models (VFMs) such as CLIP~\cite{radford2021clip},
SAM~\cite{kirillov2023sam}, and DINO~\cite{caron2021dino} achieve strong
performance on their respective tasks. Recent training-free pipelines
combine them into collaborative systems for open-vocabulary dense
prediction. Frameworks such as ProxyCLIP~\cite{lan2024proxyclip},
NACLIP~\cite{hajimiri2025naclip}, and Trident~\cite{shi2024trident}
route intermediate features from one VFM to modulate the feature
aggregation of another, and reach state-of-the-art open-vocabulary
segmentation without any annotation or fine-tuning. This paradigm has
become a default design choice for dense prediction, and it is
increasingly deployed in safety-critical applications such as
autonomous driving and medical image analysis~\cite{koleilat2024medclipsam}.

\begin{figure}[t]
\centering
\includegraphics[width=\linewidth]{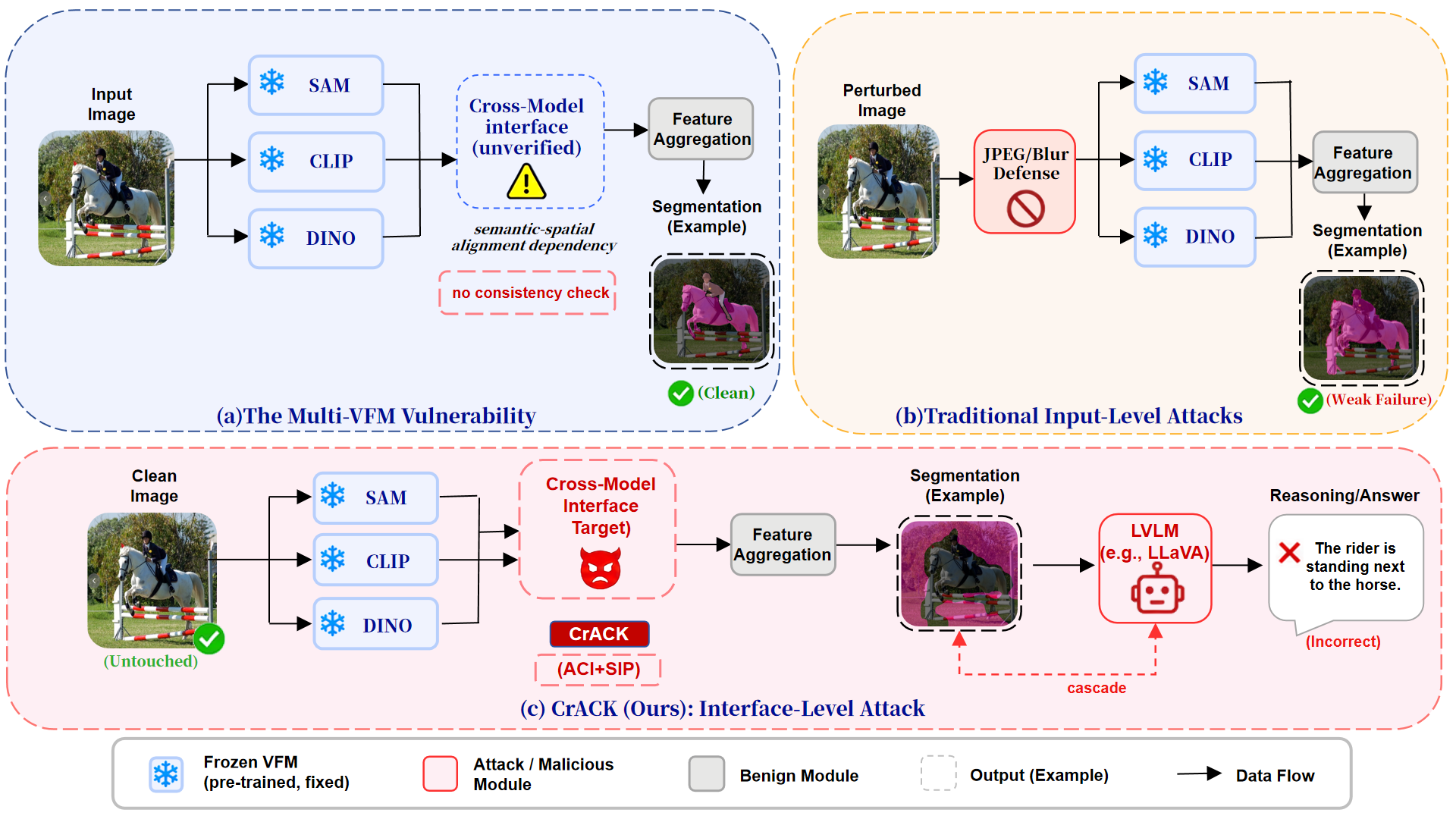}
\vspace{1mm}
\caption{\textbf{Conceptual comparison of adversarial threat paradigms.}
(a) \textbf{The Multi-VFM Vulnerability.} Collaborative systems rely
on an implicit semantic-spatial alignment dependency, leaving the
inter-model communication interface completely unverified.
(b) \textbf{Traditional Input Attacks.} Existing adversarial methods
typically target the input space by injecting human-imperceptible
noise into the raw image.
(c) \textbf{CrACK (Ours).} We introduce a new threat vector that
leaves the input image entirely pristine, and instead hijacks the
unguarded connection layer between specialized VFMs at inference
time, precipitating cascaded logical failures.}
\label{fig:threat_comparison}
\end{figure}

Despite this strong performance, collaborative VFM pipelines contain a
vulnerability that has been largely overlooked\cite{wang2026testtimeprototypeadaptationopenvocabulary}. Each model in the
pipeline consumes the intermediate output of its predecessor without
verifying semantic consistency, and the pipeline propagates these
representations across modality boundaries as if they were inherently
aligned. We refer to this unverified premise as the
\textbf{semantic-spatial alignment dependency}. It exposes the
interface between models, rather than any model itself, as the weak
point of the system. However, existing adversarial attacks have not
examined this defect, and instead perturb features within a single
model. Common attacks such as input-level attacks and backdoor attacks
each have clear limitations, as illustrated in
Figure~\ref{fig:threat_comparison}. Moreover, adversarial attacks have
been studied mainly in the LLM domain, while attacks on VFMs remain
relatively scarce. Existing VFM attacks are therefore both limited in
number and inherit the limitations of methods designed for the
single-model setting.
 
Specifically, adversarial attacks on vision models typically fall into
two categories. Input-level attacks~\cite{madry2017pgd,
goodfellow2014fgsm, zhang2023attacksam} inject pixel perturbations into
the raw image to mislead a single model, but their signal is diluted
once the perturbed features pass through a second VFM, and it is
further weakened by input-side defenses such as Gaussian smoothing and
JPEG compression. Backdoor attacks~\cite{gu2017badnets,
jia2022badencoder, bai2024badclip, xu2024shadowcast, liang2025vltrojan,
wang2025mtattack} implant a trigger by poisoning training data or
fine-tuning the victim, but they require access to a training pipeline
that training-free systems do not provide. Both categories treat the
vulnerability as residing inside a single model. In the LLM domain,
recent work shows that multi-model systems are themselves vulnerable,
since the output of one model can be manipulated to mislead
another~\cite{gu2024agentsmith}. Based on this observation, we find
that collaborative VFM pipelines share the same weakness: the
communication between models, namely the inter-model feature
interface, is itself an attack surface that no existing method targets.

Motivated by this gap, we shift the attack target from a single model
to the channel that connects them. We propose \textbf{CrACK}
(Cross-model Adversarial Consistency Attack), an attack
that operates on the inter-model interface without modifying any input
pixel, model weight, or training data. As shown in
Fig.~\ref{fig:threat_comparison}, CrACK leaves the input image
pristine and instead hijacks the unverified connection between VFMs.
CrACK turns the semantic-spatial alignment dependency against the
pipeline through two stages. Adversarial Affinity Contradiction
Injection (ACI) uses CLIP patch-level similarity to invert the affinity
matrix that the pipeline derives from SAM at the aggregation step.
Semantic Interface Poisoning (SIP) replaces the predicted labels with a
deterministic max-distance permutation built from CLIP text embeddings.
The corrupted representation is passed downstream through the same
trusted channel, so the failure cascades from segmentation into large
vision-language model reasoning, while every individual VFM output
stays unchanged.
 
In summary, our main contributions are threefold:
\begin{itemize}
  \item We propose CrACK, an inference-time attack framework against
  training-free collaborative VFM pipelines, which corrupts the
  inter-model feature interface without modifying any input pixel,
  model weight, or training data.
  \item CrACK comprises two stages, ACI and SIP. ACI inverts the
  SAM-derived affinity matrix using CLIP patch-level semantics, which
  collapses the cross-modal feature aggregation; SIP applies a
  deterministic max-distance label permutation, which steers every
  prediction toward its semantically most distant class.
  \item We validate CrACK on four collaborative VFM pipelines across
  eight segmentation datasets, demonstrating the effectiveness and the
  cross-architecture generality of the attack. Furthermore, we conduct
  experiments on a downstream VQA task, showing that the attack
  generalizes to multimodal reasoning.
\end{itemize}
 
\section{Related Work}
\label{sec:related}
 
\paragraph{Collaborative Vision Foundation Model Frameworks.}
Training-free collaborative frameworks for open-vocabulary segmentation
combine multiple VFMs, and can be classified by their cross-model
aggregation mechanism into three categories: attention-substitution
methods that modify CLIP's self-attention, DINO-prior methods that use
DINO similarity as a spatial prior, and SAM-aggregation methods that
build a global affinity from SAM. MaskCLIP~\cite{zhou2022maskclip}
belongs to the first category and replaces CLIP's last query-key
attention with a near-identity convolution to mitigate spatial
invariance. ProxyCLIP~\cite{lan2024proxyclip} and
NACLIP~\cite{hajimiri2025naclip} fall into the second category and
incorporate DINO cosine similarity as a spatial prior for CLIP feature
aggregation. Trident~\cite{shi2024trident} represents the third category
and aggregates CLIP and DINO features through an affinity matrix
derived from SAM, achieving state-of-the-art performance. The API
framework~\cite{yu2024api} further extends the collaborative chain to
generative tasks by using dense VFM feature maps as visual prompts for
LVLMs. However, all of these frameworks rely on an inter-model feature
interface that is never verified for consistency at runtime. While
multi-model systems have been shown to be adversarially vulnerable in
the LLM domain, where adversarial signals jailbreak aligned models and
cascade across collaborating agents~\cite{qi2024visual, gu2024agentsmith},
the security of the inter-VFM interface remains unexamined. CrACK
identifies this interface as a uniquely effective attack surface.
 
\paragraph{Adversarial Attacks on Vision Models.}
Adversarial attacks on vision models can be classified by their attack
surface into two categories: input-level attacks that perturb the raw
pixels of a model, and training-time attacks that implant a backdoor
through data poisoning or fine-tuning. For input-level attacks,
adversarial robustness was first studied in closed-set
segmentation~\cite{xie2017adversarial}, and subsequent work attacked
SAM with adversarial prompts~\cite{zhang2023attacksam} or corrupted
encoder representations~\cite{dong2023attacksam}. For training-time
attacks, BadNets~\cite{gu2017badnets} established the backdoor
paradigm; BadEncoder~\cite{jia2022badencoder} extended it to
pre-trained encoders that downstream systems reuse;
BadCLIP~\cite{bai2024badclip} implanted backdoors in CLIP through
trigger-aware prompt learning; VL-Trojan~\cite{liang2025vltrojan}
optimized multimodal triggers for autoregressive LVLMs; and
ShadowCast~\cite{xu2024shadowcast} performed stealthy data poisoning
against VLMs, with recent work showing that web-scale poisoning is
practical~\cite{carlini2024poisoning}. However, input-level attacks
keep their adversarial signal in the input image, so it is diluted once
the perturbed features pass through a second VFM and is filtered by
input-side defenses, while training-time attacks require access to a
training pipeline that training-free systems do not expose. Both
categories locate the vulnerability inside a single model and can, in
principle, be mitigated by hardening that model alone. None targets the
feature interface between models, which is precisely the surface CrACK
exploits.

\section{Method}
\label{sec:method}
 
\subsection{Threat Model}
\label{sec:threat}
 
\paragraph{Problem formulation.}
A collaborative VFM pipeline produces a dense prediction by combining
several frozen VFMs. We denote it as
$f_{\mathrm{pipe}}(x;\theta_{\mathrm{CLIP}},\theta_{\mathrm{SAM}},
\theta_{\mathrm{DINO}})$, where $x$ is the input image and $\theta_*$
are publicly released and frozen weights. The pipeline extracts
features from each VFM in parallel, fuses them at an aggregation step,
and matches the fused feature against CLIP text embeddings to produce a
segmentation map. The aggregation step is the inter-model interface
that CrACK targets.
 
\paragraph{Victim pipeline.}
We instantiate the victim pipeline with Trident~\cite{shi2024trident},
which uses the affinity aggregation interface shared by
SCLIP~\cite{wang2023sclip}, ProxyCLIP~\cite{lan2024proxyclip}, and
NACLIP~\cite{hajimiri2025naclip}. Trident fuses CLIP and DINO features
through an affinity matrix $A$ derived from SAM. Let $F$ denote the SAM
encoder features and $W$ the head-averaged attention weights of SAM's
last encoder layer. Trident builds $A$ as
\begin{equation}
  C_{ij}=\cos(F_i,F_j),\qquad
  A=\mathrm{Norm}\!\left(W\odot\mathbf{1}[C\ge\epsilon]\right),
  \label{eq:trident_aff}
\end{equation}
where $\epsilon$ is a data-adaptive threshold, $\odot$ is the
element-wise product, $\mathbf{1}[\cdot]$ is the indicator function,
and $\mathrm{Norm}$ is row normalization. The matrix $A$ reweights the
spliced CLIP feature map $I_{\mathrm{feat}}$ into the aggregated
feature $\tilde{I}_{\mathrm{feat}}=A\cdot I_{\mathrm{feat}}$ used for
segmentation. Equation~\ref{eq:trident_aff} encodes the
semantic-spatial alignment dependency: it assumes that patch pairs SAM
finds similar also deserve coherent semantic aggregation for CLIP.
CrACK attacks this assumption.
 
\paragraph{Attacker capability.}
We adopt an inference-time threat model. The attacker injects
lightweight perturbation logic into the aggregation glue code that
connects the frozen VFMs, and modifies no VFM weight, no VFM forward
function, and no input pixel. This capability is realistic in two
deployment settings. In inference-time supply-chain poisoning,
training-free pipelines are distributed as thin aggregation
repositories that import third-party encoders, and a backdoor planted
in this layer needs no access to the encoders~\cite{gu2017badnets,
carlini2024poisoning}. In microservice API hijacking, VFMs are served
as separate microservices, and the aggregation step becomes a network
exchange of feature tensors that an adversary on the path can rewrite.
To compute the perturbation, the attacker uses one public
CLIP-ViT-B/16 instance, whose forward pass is identical to any
independent copy and need not run inside the victim trust domain.
 
\paragraph{Stealthiness.}
The attack lives only in the aggregation step. Every VFM therefore
produces its exact clean standalone output, and the input image stays
unchanged. The attack is invisible to input-side preprocessing,
per-model output inspection, and weight-integrity verification. Its
stealthiness is structural rather than perturbative.

\begin{figure}[t]
    \includegraphics[width=\linewidth]{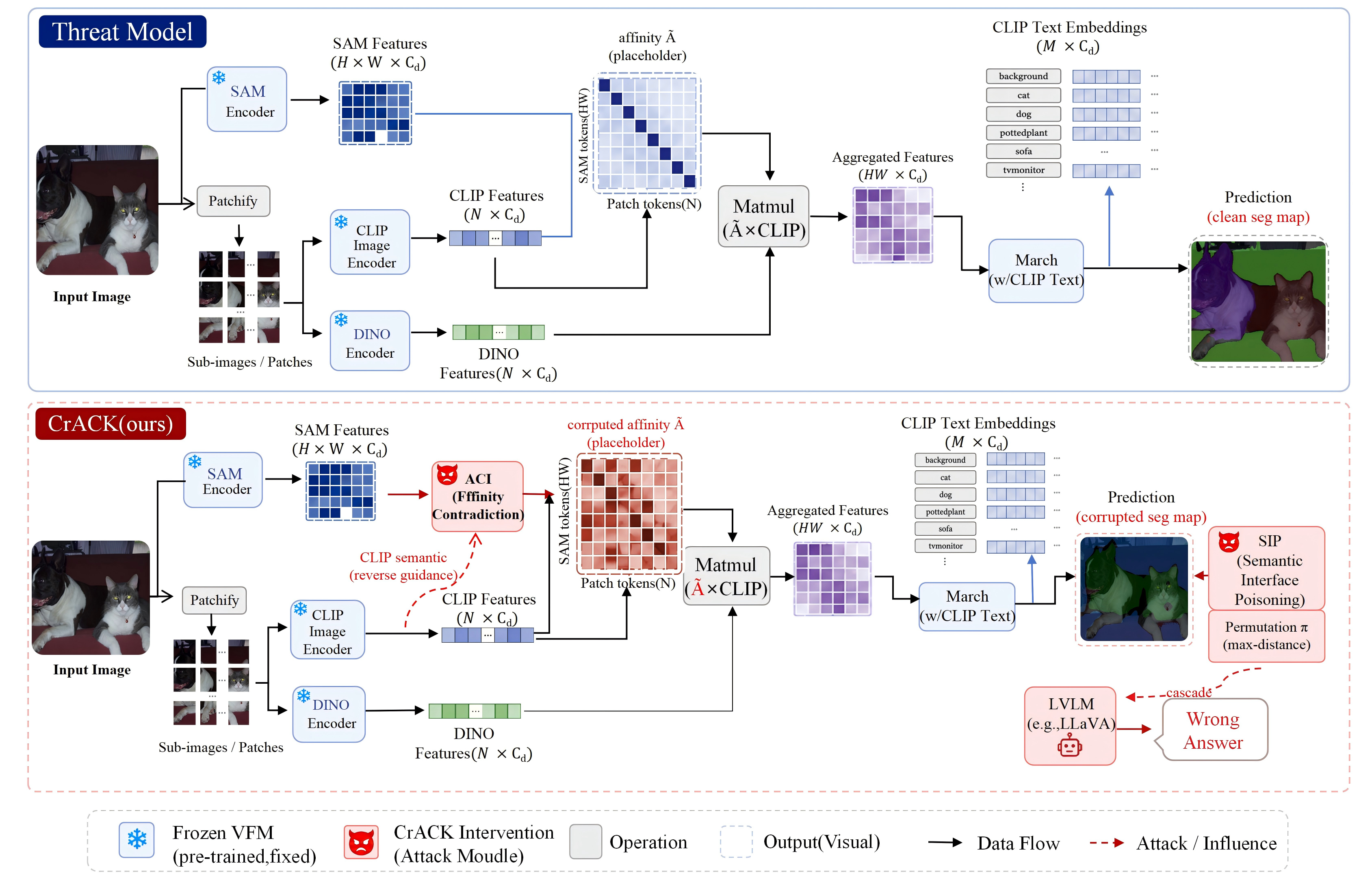}
    \vspace{1mm}
    \caption{\textbf{Overview of CrACK.} The top row shows the benign
    collaborative pipeline instantiated on Trident, where SAM, CLIP, and
    DINO branches are processed in parallel and fused through an affinity
    matrix $A$ into a clean segmentation map. The bottom row shows CrACK,
    which keeps every model frozen and intervenes at two points.
    Adversarial Affinity Contradiction Injection (ACI) uses CLIP
    patch-level semantics to invert the SAM-derived affinity, turning $A$
    into the corrupted $\tilde{A}$. Semantic Interface Poisoning (SIP)
    applies a deterministic max-distance label permutation $\pi^{*}$ at the
    prediction interface. The corrupted dense feature is forwarded, through
    the same trusted channel, into a downstream LVLM, where the
    interface-level contradiction cascades into erroneous reasoning.}
    \label{fig:crack_architecture}
    \centering
\end{figure}

\subsection{The Proposed CrACK}
\label{sec:overview}
 
A collaborative VFM pipeline exposes two trust points: the affinity
aggregation step, where cross-model features are fused, and the
prediction interface, where the label map is handed to downstream
modules. The overall architecture is depicted in Fig.~\ref{fig:crack_architecture}.
We present CrACK, an inference-time attack framework that
attacks both. CrACK operates in two stages. Stage~1, Adversarial
Affinity Contradiction Injection (ACI), corrupts the affinity matrix at
the feature level so that the aggregation step fuses semantically
contradictory regions. Stage~2, Semantic Interface Poisoning (SIP),
poisons the prediction interface at the label level so that the
corrupted segmentation map steers downstream reasoning. The two stages
form a single inference-time forward pass and are described below.
 
\subsection{Adversarial Affinity Contradiction Injection}
\label{sec:aci}

The affinity matrix $A$ in Eq.~\ref{eq:trident_aff} determines which
patches are aggregated together, and it is built from SAM under the
assumption that visually similar patches deserve coherent semantic
treatment. To turn this assumption into a failure mode, we propose
Adversarial Affinity Contradiction Injection (ACI), which recombines
the SAM representation so that the affinity gates in semantically
contradictory patch pairs instead of consistent ones. ACI takes the
SAM features $F$ and attention $W$ as input and outputs a corrupted
affinity matrix $\tilde{A}$ that replaces $A$.
 
\paragraph{CLIP-guided inverse similarity.}
ACI needs a reliable signal of which patch pairs are semantically
dissimilar, so that it knows the adversarial aggregation direction. We
obtain this signal from CLIP itself. We extract CLIP patch tokens
$\{t_i\}$ from the input image with the frozen CLIP visual encoder, and
compute the semantic similarity matrix
\begin{equation}
  S_{ij}=\cos(t_i,t_j),\qquad
  \bar{S}_{ij}=\frac{S_{ij}-S_{\min}}{S_{\max}-S_{\min}},
  \label{eq:clip_sim}
\end{equation}
where $t_i$ is the $L_2$-normalized CLIP patch token at position $i$.
We define the inverse similarity $\hat{S}=1-\bar{S}$, whose entry
$\hat{S}_{ij}$ is large for patch pairs CLIP regards as semantically
dissimilar and small for intra-class pairs. We row-normalize $\hat{S}$
into mixing weights $\hat{S}^{\mathrm{row}}$.
 
\paragraph{Feature and attention inversion.}
The affinity in Eq.~\ref{eq:trident_aff} is built from two inputs, the
feature-based mask $\mathbf{1}[C\ge\epsilon]$ and the attention weights
$W$. Corrupting only one of them is insufficient: inverting the feature
alone leaves the clean $W$ partially correct, and inverting the
attention alone is suppressed by the clean mask. To corrupt both, ACI
first replaces each SAM feature with a mixture of its semantically
dissimilar counterparts,
\begin{equation}
  \tilde{F}_i=\sum_j\hat{S}^{\mathrm{row}}_{ij}\,F_j,
  \label{eq:feat_inv}
\end{equation}
which drives originally similar patches apart and originally dissimilar
patches together, so that the cosine matrix $\tilde{C}$ computed on
$\tilde{F}$ approximately inverts the clean $C$. ACI then replaces the
attention weights with the inverse similarity rescaled to the original
magnitude,
\begin{equation}
  \tilde{W}_{ij}=\hat{S}_{ij}\cdot\overline{|W|},
  \label{eq:attn_inv}
\end{equation}
where $\overline{|W|}$ is the mean absolute attention. The corrupted
affinity is then
\begin{equation}
  \tilde{A}=\mathrm{Norm}\!\left(\tilde{W}\odot
  \mathbf{1}[\tilde{C}\ge\epsilon]\right).
  \label{eq:corrupt_aff}
\end{equation}
After aggregation, each foreground feature becomes a weighted average
of background features, so the downstream argmax assigns systematically
incorrect labels, while every VFM standalone output stays intact.
Algorithm~\ref{alg:aci} summarizes the procedure. While Eqs.~\ref{eq:clip_sim}--\ref{eq:corrupt_aff} instantiate ACI on SAM, the same inverse-similarity mixing applies to any VFM backbone; for DINO-based pipelines such as ProxyCLIP and NACLIP, ACI substitutes the DINO intermediate features in place of SAM features.
 
\begin{algorithm}[t]
\caption{Adversarial Affinity Contradiction Injection (ACI)}
\label{alg:aci}
\begin{algorithmic}[1]
\Require SAM features $F$, SAM attention $W$, input image $x$,
         threshold $\epsilon$
\Ensure Corrupted affinity matrix $\tilde{A}$
  \State $\{t_i\}\gets\mathrm{CLIP}_{\mathrm{img}}(x)$
         \Comment{patch tokens from frozen CLIP}
  \State $S_{ij}\gets\cos(t_i,t_j)$;\quad
         $\bar{S}\gets\mathrm{MinMax}(S)$
         \Comment{Eq.~\ref{eq:clip_sim}}
  \State $\hat{S}\gets 1-\bar{S}$;\quad
         $\hat{S}^{\mathrm{row}}\gets\mathrm{RowNorm}(\hat{S})$
         \Comment{inverse similarity}
  \State $\tilde{F}_i\gets\sum_j\hat{S}^{\mathrm{row}}_{ij}F_j$
         \Comment{Eq.~\ref{eq:feat_inv}}
  \State $\tilde{W}_{ij}\gets\hat{S}_{ij}\cdot\overline{|W|}$
         \Comment{Eq.~\ref{eq:attn_inv}}
  \State $\tilde{C}_{ij}\gets\cos(\tilde{F}_i,\tilde{F}_j)$
  \State $\tilde{A}\gets\mathrm{RowNorm}\!\left(\tilde{W}\odot
         \mathbf{1}[\tilde{C}\ge\epsilon]\right)$
         \Comment{Eq.~\ref{eq:corrupt_aff}}
  \State \Return $\tilde{A}$
\end{algorithmic}
\end{algorithm}
 
\subsection{Semantic Interface Poisoning}
\label{sec:sip}
 
ACI corrupts the affinity at the feature level, but the prediction
interface, namely the label map that downstream modules consume, is
read verbatim without any semantic verification. To exploit this second
trust point, we propose Semantic Interface Poisoning (SIP), which
replaces the predicted labels with a deterministic permutation that
maximizes semantic contradiction. SIP takes the predicted label map
$\hat{Y}$ as input and outputs a poisoned label map $\tilde{Y}$, using
only CLIP text-embedding geometry and no gradient optimization.
 
\paragraph{Max-distance permutation.}
To make every substituted label as misleading as possible, SIP assigns
each class the semantically most distant target in the vocabulary.
Given the class vocabulary $\mathcal{C}$ and the $L_2$-normalized CLIP
text embeddings $\{\mathbf{e}_c\}$, we compute the cross-class
dissimilarity
\begin{equation}
  D_{c,c'}=1-\cos(\mathbf{e}_c,\mathbf{e}_{c'}),\qquad c\neq c'.
  \label{eq:sip_dist}
\end{equation}
We then build a bijective permutation $\pi^{*}$ that greedily assigns
to each class the most distant target not yet used,
\begin{equation}
  \pi^{*}(c)=\argmax_{c'\in\mathcal{C}\setminus\mathcal{U}}D_{c,c'},
  \qquad \mathcal{U}\leftarrow\mathcal{U}\cup\{\pi^{*}(c)\},
  \label{eq:sip_perm}
\end{equation}
where $\mathcal{U}$ accumulates already-assigned targets to keep
$\pi^{*}$ bijective. Each substituted label therefore lies in the
region of CLIP embedding space antipodal to the correct class, which is
the maximum contradiction expressible within the closed label set.
Algorithm~\ref{alg:sip} gives the full construction.
 
\paragraph{Prediction poisoning and cascaded effect.}
SIP applies $\pi^{*}$ pixel-wise at the prediction interface,
\begin{equation}
  \tilde{Y}_i=\pi^{*}(\hat{Y}_i),\qquad
  \forall\, i\in\{1,\ldots,HW\},
  \label{eq:sip_apply}
\end{equation}
with no additional compute. The output $\tilde{Y}$ is a syntactically
valid segmentation map, so SIP is transparent to any checker that
inspects prediction format. When SIP follows ACI, the complete CrACK
pipeline is
\begin{equation}
  \tilde{Y}_i=\pi^{*}\!\left(\argmax_c\;
  \bigl[\tilde{A}\cdot I_{\mathrm{feat}}\bigr]_i\,
  \mathbf{e}_c^{\top}\right).
  \label{eq:full_crack}
\end{equation}
ACI shifts the pre-permutation logits toward incorrect classes, and SIP
redirects each remaining correct prediction to its antipodal class. The
corrupted dense feature, forwarded through the same trusted channel,
propagates the failure into downstream LVLM reasoning, where the
interface-level contradiction re-emerges as an erroneous response.
 
\begin{algorithm}[t]
\caption{Semantic Interface Poisoning (SIP)}
\label{alg:sip}
\begin{algorithmic}[1]
\Require Class vocabulary $\mathcal{C}$, CLIP text embeddings
         $\{\mathbf{e}_c\}$, predicted label map $\hat{Y}$
\Ensure Poisoned label map $\tilde{Y}$
  \State $D_{c,c'}\gets 1-\cos(\mathbf{e}_c,\mathbf{e}_{c'})$
         \Comment{Eq.~\ref{eq:sip_dist}}
  \State $\mathcal{U}\gets\emptyset$
  \For{each class $c\in\mathcal{C}$}
    \State $\pi^{*}(c)\gets\argmax_{c'\in\mathcal{C}\setminus\mathcal{U}}
           D_{c,c'}$ \Comment{most distant free target}
    \State $\mathcal{U}\gets\mathcal{U}\cup\{\pi^{*}(c)\}$
  \EndFor
  \State $\tilde{Y}_i\gets\pi^{*}(\hat{Y}_i)$ for every pixel $i$
         \Comment{Eq.~\ref{eq:sip_apply}}
  \State \Return $\tilde{Y}$
\end{algorithmic}
\end{algorithm}

\section{Experiments}
\label{sec:exp}

\subsection{Experimental Setup}
\label{sec:expsetup}

\paragraph{Datasets and victim pipelines.}
Following Trident~\cite{shi2024trident}, we evaluate CrACK on eight
open-vocabulary segmentation benchmarks: VOC20 and
VOC21~\cite{everingham2010pascal}, Context59 and
Context60~\cite{mottaghi2014role}, COCO Object~\cite{lin2014microsoft},
COCO Stuff~\cite{caesar2018coco},
Cityscapes~\cite{cordts2016cityscapes}, and
ADE20k~\cite{zhou2019semantic}. We attack four representative
training-free pipelines that span three aggregation paradigms:
SCLIP~\cite{wang2023sclip}, which substitutes the CLIP self-attention;
ProxyCLIP~\cite{lan2024proxyclip} and
NACLIP~\cite{hajimiri2025naclip}, which use DINO similarity as a
spatial prior; and Trident~\cite{shi2024trident}, which aggregates
features through a SAM-derived affinity matrix. All pipelines use the
CLIP ViT-B/16 backbone. For downstream evaluation, we use
LLaVA-1.5-13B~\cite{liu2023llava} as the reasoning engine and follow
the API visual prompting protocol~\cite{yu2024api}, matching the
configuration adopted in Trident~\cite{shi2024trident} for fair
comparison.

\paragraph{Baselines.}
We compare CrACK with three input-level adversarial attacks:
FGSM~\cite{goodfellow2014fgsm}, PGD-SAM, and PGD-CLIP. PGD-SAM and
PGD-CLIP optimize a 10-step PGD perturbation on the SAM encoder feature
loss and the CLIP patch feature loss respectively, with budget
$\varepsilon{=}8/255$ and step size $\alpha{=}1/255$. All perturbed
inputs are forwarded through the complete victim pipeline, so every
method is measured on the same task. These baselines share the same
victim pipeline as CrACK but differ in the attack surface: they perturb
input pixels, while CrACK acts at the aggregation interface.

\paragraph{Metrics.}
We report mean Intersection over Union (mIoU) for segmentation, where
lower values indicate stronger attacks. We define attack success rate
(ASR) as the fraction of foreground pixels that are correctly predicted
by the clean pipeline but flipped after the attack:
\begin{equation}
  \mathrm{ASR}=
  \frac{\bigl|\{i\,\vert\, \hat{y}^{\mathrm{cl}}_i = y_i \,\wedge\,
  \tilde{y}_i \neq y_i\}\bigr|}
       {\bigl|\{i\,\vert\, \hat{y}^{\mathrm{cl}}_i = y_i\}\bigr|},
  \label{eq:asr}
\end{equation}
where $\hat{y}^{\mathrm{cl}}_i$ and $\tilde{y}_i$ denote the clean and
attacked predictions at pixel $i$, and $y_i$ is the ground truth. For
downstream evaluation, we report accuracy on POPE~\cite{li2023pope}
and GPT-4o judge scores on MM-Vet~\cite{yu2024mmvet} and
LLaVA-Wild~\cite{liu2023llava}.

\paragraph{Implementation.}
CLIP patch tokens are extracted from the penultimate transformer block.
The max-distance permutation $\pi^{*}$ is precomputed once per dataset
vocabulary and reused at inference without optimization. CrACK uses the
same hyperparameters across all four victim pipelines without
per-pipeline tuning. All experiments run on a single NVIDIA H100 GPU.

\subsection{Attack Effectiveness on Open-Vocabulary Segmentation}
\label{sec:seg_res}

Table~\ref{tab:main} reports segmentation performance under CrACK and
the three baseline attacks on four victim pipelines across eight
benchmarks. We have three observations.

First, CrACK produces the strongest degradation on every pipeline and
every dataset, reducing the average mIoU of Trident from $45.8$ to
$1.55$ and showing the same collapse on SCLIP, ProxyCLIP, and NACLIP.
Second, the same collapse occurs across all three aggregation paradigms
represented by these pipelines, indicating that the vulnerability
resides in the cross-model aggregation interface shared by these
pipelines rather than in any specific aggregation mechanism. This
generality is particularly evident on NACLIP, where CrACK reduces the
average mIoU from $42.5$ to $0.27$ with an ASR of $98.0\%$, despite
NACLIP using a DINO-based spatial prior rather than the SAM-derived
affinity targeted by ACI. Third, the input-level baselines are
substantially weaker. PGD-SAM and PGD-CLIP perturb input pixels and
influence only a single branch of the aggregation step. As reported in
Table~\ref{tab:main}, the average mIoU drops by $0.99$, $1.20$,
$1.17$, and $1.51$ on SCLIP, ProxyCLIP, NACLIP, and Trident
respectively, with the average ASR reaching at most $4.06\%$ across
pipelines. On Trident with VOC21, for example, PGD-CLIP reduces mIoU
only to $64.45$, leaving the segmentation essentially intact, whereas
CrACK reduces it to $2.88$. CrACK corrupts both the affinity branch
through ACI and the prediction interface through SIP, yielding a much
larger gap. These results support our central claim: the
semantic-spatial alignment dependency is the structural vulnerability
of collaborative VFM pipelines, and attacking it directly is more
effective than perturbing any single model.

\begin{table}[t]
\centering
  \caption{Effectiveness of CrACK against four training-free
  open-vocabulary segmentation pipelines across eight benchmarks with
  CLIP ViT-B/16. None reports the clean baseline. The PGD column
  reports PGD-CLIP under the same victim pipeline; PGD-SAM yields
  similar values and is omitted for brevity. Lower mIoU and higher
  ASR indicate stronger attacks.}
  \vspace{1.5mm}
  \label{tab:main}
  \scriptsize
  \resizebox{\textwidth}{!}{%
  \setlength{\tabcolsep}{2.4pt}
  \begin{tabular}{l l ccc ccc ccc ccc}
    \toprule
    \multirow{2}{*}{Dataset} & \multirow{2}{*}{Metric}
      & \multicolumn{3}{c}{SCLIP}
      & \multicolumn{3}{c}{ProxyCLIP}
      & \multicolumn{3}{c}{NACLIP}
      & \multicolumn{3}{c}{Trident} \\
    \cmidrule(lr){3-5}\cmidrule(lr){6-8}\cmidrule(lr){9-11}\cmidrule(lr){12-14}
    & & None & PGD & \cellcolor{ourshl}\textbf{CrACK}
        & None & PGD & \cellcolor{ourshl}\textbf{CrACK}
        & None & PGD & \cellcolor{ourshl}\textbf{CrACK}
        & None & PGD & \cellcolor{ourshl}\textbf{CrACK} \\
    \midrule
    \multirow{2}{*}{VOC21}
      & mIoU $\downarrow$
        & 61.7 & 60.47 & \cellcolor{ourshl}\textbf{3.50}
        & 61.3 & 59.77 & \cellcolor{ourshl}\textbf{3.56}
        & 64.1 & 62.50 & \cellcolor{ourshl}\textbf{1.19}
        & 67.1 & 64.45 & \cellcolor{ourshl}\textbf{2.88} \\
      & ASR (\%) $\uparrow$
        & -- & 2.12 & \cellcolor{ourshl}\textbf{100.0}
        & -- & 2.74 & \cellcolor{ourshl}\textbf{100.0}
        & -- & 2.58 & \cellcolor{ourshl}\textbf{94.1}
        & -- & 3.81 & \cellcolor{ourshl}\textbf{99.8} \\
    \multirow{2}{*}{Context60}
      & mIoU $\downarrow$
        & 31.5 & 30.71 & \cellcolor{ourshl}\textbf{0.01}
        & 35.3 & 34.24 & \cellcolor{ourshl}\textbf{0.00}
        & 35.0 & 33.95 & \cellcolor{ourshl}\textbf{0.09}
        & 38.6 & 37.18 & \cellcolor{ourshl}\textbf{0.06} \\
      & ASR (\%) $\uparrow$
        & -- & 2.45 & \cellcolor{ourshl}\textbf{100.0}
        & -- & 3.12 & \cellcolor{ourshl}\textbf{100.0}
        & -- & 2.97 & \cellcolor{ourshl}\textbf{99.9}
        & -- & 4.15 & \cellcolor{ourshl}\textbf{98.3} \\
    \multirow{2}{*}{Object}
      & mIoU $\downarrow$
        & 32.1 & 30.82 & \cellcolor{ourshl}\textbf{0.86}
        & 37.5 & 36.19 & \cellcolor{ourshl}\textbf{0.87}
        & 36.2 & 34.93 & \cellcolor{ourshl}\textbf{0.03}
        & 41.1 & 39.4  & \cellcolor{ourshl}\textbf{0.81} \\
      & ASR (\%) $\uparrow$
        & -- & 2.78 & \cellcolor{ourshl}\textbf{100.0}
        & -- & 3.45 & \cellcolor{ourshl}\textbf{100.0}
        & -- & 3.21 & \cellcolor{ourshl}\textbf{99.0}
        & -- & 4.62 & \cellcolor{ourshl}\textbf{98.3} \\
    \multirow{2}{*}{VOC20}
      & mIoU $\downarrow$
        & 83.5 & 82.67 & \cellcolor{ourshl}\textbf{1.17}
        & 80.3 & 79.10 & \cellcolor{ourshl}\textbf{0.15}
        & 83.0 & 81.76 & \cellcolor{ourshl}\textbf{0.74}
        & 84.5 & 83.7  & \cellcolor{ourshl}\textbf{4.05} \\
      & ASR (\%) $\uparrow$
        & -- & 1.24 & \cellcolor{ourshl}\textbf{98.6}
        & -- & 1.68 & \cellcolor{ourshl}\textbf{100.0}
        & -- & 1.45 & \cellcolor{ourshl}\textbf{90.7}
        & -- & 1.73 & \cellcolor{ourshl}\textbf{95.8} \\
    \multirow{2}{*}{Context59}
      & mIoU $\downarrow$
        & 36.1 & 35.02 & \cellcolor{ourshl}\textbf{0.19}
        & 39.1 & 37.73 & \cellcolor{ourshl}\textbf{0.02}
        & 38.4 & 37.25 & \cellcolor{ourshl}\textbf{0.09}
        & 42.2 & 40.67 & \cellcolor{ourshl}\textbf{0.13} \\
      & ASR (\%) $\uparrow$
        & -- & 2.95 & \cellcolor{ourshl}\textbf{100.0}
        & -- & 3.34 & \cellcolor{ourshl}\textbf{100.0}
        & -- & 3.18 & \cellcolor{ourshl}\textbf{100.0}
        & -- & 3.95 & \cellcolor{ourshl}\textbf{98.7} \\
    \multirow{2}{*}{Stuff}
      & mIoU $\downarrow$
        & 23.9 & 23.18 & \cellcolor{ourshl}\textbf{0.09}
        & 26.5 & 25.71 & \cellcolor{ourshl}\textbf{0.05}
        & 25.7 & 24.93 & \cellcolor{ourshl}\textbf{0.00}
        & 28.3 & 27.6  & \cellcolor{ourshl}\textbf{0.03} \\
      & ASR (\%) $\uparrow$
        & -- & 2.43 & \cellcolor{ourshl}\textbf{100.0}
        & -- & 2.96 & \cellcolor{ourshl}\textbf{100.0}
        & -- & 2.82 & \cellcolor{ourshl}\textbf{100.0}
        & -- & 3.28 & \cellcolor{ourshl}\textbf{98.3} \\
    \multirow{2}{*}{City}
      & mIoU $\downarrow$
        & 34.1 & 32.74 & \cellcolor{ourshl}\textbf{2.09}
        & 38.1 & 36.58 & \cellcolor{ourshl}\textbf{1.98}
        & 38.3 & 36.77 & \cellcolor{ourshl}\textbf{0.04}
        & 42.9 & 40.4  & \cellcolor{ourshl}\textbf{4.37} \\
      & ASR (\%) $\uparrow$
        & -- & 3.21 & \cellcolor{ourshl}\textbf{100.0}
        & -- & 3.87 & \cellcolor{ourshl}\textbf{100.0}
        & -- & 3.65 & \cellcolor{ourshl}\textbf{99.9}
        & -- & 4.88 & \cellcolor{ourshl}\textbf{98.3} \\
    \multirow{2}{*}{ADE}
      & mIoU $\downarrow$
        & 17.8 & 17.27 & \cellcolor{ourshl}\textbf{0.16}
        & 20.2 & 19.49 & \cellcolor{ourshl}\textbf{0.12}
        & 19.1 & 18.53 & \cellcolor{ourshl}\textbf{0.00}
        & 21.9 & 20.9  & \cellcolor{ourshl}\textbf{0.03} \\
      & ASR (\%) $\uparrow$
        & -- & 2.87 & \cellcolor{ourshl}\textbf{100.0}
        & -- & 3.41 & \cellcolor{ourshl}\textbf{100.0}
        & -- & 3.27 & \cellcolor{ourshl}\textbf{99.9}
        & -- & 6.04 & \cellcolor{ourshl}\textbf{98.3} \\
    \midrule
    \multirow{2}{*}{Avg.}
      & mIoU $\downarrow$
        & 40.1 & 39.11 & \cellcolor{ourshl}\textbf{1.01}
        & 42.3 & 41.10 & \cellcolor{ourshl}\textbf{0.84}
        & 42.5 & 41.33 & \cellcolor{ourshl}\textbf{0.27}
        & 45.8 & 44.29 & \cellcolor{ourshl}\textbf{1.55} \\
      & ASR (\%) $\uparrow$
        & -- & 2.51 & \cellcolor{ourshl}\textbf{99.8}
        & -- & 3.07 & \cellcolor{ourshl}\textbf{100.0}
        & -- & 2.89 & \cellcolor{ourshl}\textbf{98.0}
        & -- & 4.06 & \cellcolor{ourshl}\textbf{98.2} \\
    \bottomrule
  \end{tabular}}
\end{table}

\subsection{Downstream LVLM Attack via Poisoned Visual Prompts}
\label{sec:lvlm}

CrACK is designed to propagate the corruption at the SAM--CLIP interface
into the downstream reasoning engine. We follow the API visual
prompting protocol~\cite{yu2024api}, in which the dense feature map of
the segmentation pipeline modulates the input image of an LVLM. We
replace the clean dense feature map with the CrACK-corrupted variant
and feed the resulting visual prompt into LLaVA-1.5-13B.
Table~\ref{tab:lvlm} reports three settings: the LLaVA-1.5-13B
baseline, LLaVA with clean Trident features under the API protocol,
and LLaVA with CrACK-corrupted features. The corrupted visual prompt
degrades response quality on all three benchmarks, with POPE accuracy
dropping from $87.4$ to $58.3$, MM-Vet from $36.0$ to $19.4$, and
LLaVA-Wild from $73.4$ to $38.7$. All three CrACK rows lie well below
the LLaVA-only baseline ($86.9$, $31.7$, $66.9$), confirming that
CrACK does not merely neutralize the helpful Trident prompt but
actively misleads the downstream reasoning. This shows that the
contradiction injected at the segmentation stage is amplified rather
than absorbed by the LVLM. These results confirm the cascading nature
of CrACK: an attack crafted at the inter-VFM interface compromises
downstream multimodal reasoning without modifying the input image or
any model weight.

\begin{table}[t]
  \caption{Downstream LVLM degradation under CrACK-corrupted visual
  prompts on LLaVA-1.5-13B, following the API visual prompting
  protocol~\cite{yu2024api}. The dense feature map of Trident is
  replaced with the CrACK-corrupted variant before being injected as
  the visual prompt. Baseline numbers for LLaVA-1.5-13B and
  LLaVA+Trident are taken from Trident~\cite{shi2024trident}. Lower
  scores indicate stronger attacks.}
  \vspace{1.5mm}
  \label{tab:lvlm}
  \centering
  \small
  \begin{tabular}{lccc}
    \toprule
    Method & POPE & MM-Vet & LLaVA-Wild \\
    \midrule
    LLaVA-1.5-13B                            & 86.9 & 31.7 & 66.9 \\
    LLaVA + Trident (API)                    & 87.4 & 36.0 & 73.4 \\
    \rowcolor{ourshl}
    LLaVA + Trident + \textbf{CrACK (Ours)}  & \textbf{58.3} & \textbf{19.4} & \textbf{38.7} \\
    \bottomrule
  \end{tabular}
\end{table}

\subsection{Attack Stealthiness}
\label{sec:stealth}

CrACK leaves the input image and all VFM weights unchanged. Every
individual VFM produces its exact clean standalone output, so the
attack is invisible to input-side preprocessing, per-model output
inspection, and weight-integrity verification. Its stealthiness is
therefore structural rather than perturbative, and cannot be reduced to
a single perturbation-magnitude metric. We further quantify this
property by contrasting CrACK with PGD-CLIP under three input-level
defenses commonly used to mitigate adversarial perturbations: Gaussian
blur with kernel size $5$ and $\sigma{=}1.0$, JPEG compression at
quality factor $75$, and random cropping with crop ratio $0.875$
followed by resizing to the original input size. All experiments are
conducted on Trident with PASCAL VOC21.

Table~\ref{tab:stealth} reveals two contrasting patterns. First, the
three defenses do mitigate PGD-CLIP: under Blur, PGD-CLIP only reduces
mIoU from the defended clean baseline of $65.0$ to $64.62$, an attack
residual of $0.38$ mIoU; under JPEG and Crop the residual is similarly
small, and PGD-CLIP's ASR drops to below $2\%$ in all three cases. This
confirms that input-level defenses are effective against input-level
attacks, which is the historical justification for this defense
paradigm. Second, the same defenses are completely ineffective against
CrACK. The attacked mIoU stays within a $0.06$ window across all four
settings ($2.88$, $2.86$, $2.87$, $2.82$), and the ASR remains above
$98.9\%$. The most revealing case is Crop, where the defense itself
drops clean mIoU by $13.7$ -- a cost larger than the entire PGD-CLIP
attack ($2.65$ mIoU) -- yet CrACK still drives attacked mIoU down to
$2.82$. No input-side defense can simultaneously preserve clean utility
and block CrACK, because CrACK does not live in the input.

\begin{table}[t]
  \caption{Performance on Trident with PASCAL VOC21 under no defense
  and three input-level defenses. Clean mIoU is computed on the
  defended input without attack; Atk mIoU and ASR are computed on the
  defended input after attack. CrACK remains effective across all
  defenses, while PGD-CLIP is neutralized by every defense at a cost
  to clean utility.}
  \vspace{1.5mm}
  \label{tab:stealth}
  \centering
  \small
  \setlength{\tabcolsep}{5pt}
  \begin{tabular}{lccccc}
    \toprule
    \multirow{2}{*}{Defense} & \multirow{2}{*}{Clean mIoU}
      & \multicolumn{2}{c}{PGD-CLIP}
      & \multicolumn{2}{c}{\cellcolor{ourshl}\textbf{CrACK (Ours)}} \\
    \cmidrule(lr){3-4}\cmidrule(lr){5-6}
    & & Atk mIoU $\downarrow$ & ASR (\%) $\uparrow$
      & \cellcolor{ourshl}Atk mIoU $\downarrow$
      & \cellcolor{ourshl}ASR (\%) $\uparrow$ \\
    \midrule
    None & 67.1 & 64.45 & 3.81
         & \cellcolor{ourshl}\textbf{2.88} & \cellcolor{ourshl}\textbf{99.8} \\
    Blur & 65.0 & 64.62 & 1.92
         & \cellcolor{ourshl}\textbf{2.86} & \cellcolor{ourshl}\textbf{99.5} \\
    JPEG & 66.8 & 64.51 & 1.64
         & \cellcolor{ourshl}\textbf{2.87} & \cellcolor{ourshl}\textbf{99.6} \\
    Crop & 53.4 & 51.27 & 1.18
         & \cellcolor{ourshl}\textbf{2.82} & \cellcolor{ourshl}\textbf{98.9} \\
    \bottomrule
  \end{tabular}
\end{table}

\subsection{Ablation Study}
\label{sec:ablation}

Table~\ref{tab:ablation} decomposes CrACK into its two attack stages on
Trident with PASCAL VOC21. ACI alone corrupts the feature-level
affinity and reduces mIoU from $67.1$ to $15.0$ with an ASR of
$66.76\%$. SIP alone acts on the prediction interface through the
max-distance permutation and flips every correctly classified
foreground pixel, giving an ASR of $100.00\%$. The full pipeline
combines both stages and yields the strongest degradation, with an
mIoU of $2.8$ and an ASR of $100.00\%$. ACI and SIP are not redundant
components but two complementary attack stages, each targeting a
distinct trust point: ACI shifts the pre-permutation logits toward
incorrect classes at the feature level, and SIP redirects each
remaining correct prediction to its semantically most distant label
at the label level.

We further compare ACI and SIP with three trivial baselines that
exploit the same threat model---write access to the aggregation glue
code---but use unstructured destruction. $A{=}0$ replaces the affinity
with an all-zero matrix, $A{=}\mathrm{random}$ replaces it with a
row-normalized random matrix, and $Y{=}\mathrm{random\ derangement}$
replaces the predicted label map with a random class bijection
without CLIP semantic guidance. The three baselines reduce mIoU to
$3.49$, $24.32$, and $4.14$, with corresponding ASR of $99.60\%$,
$58.20\%$, and $100.00\%$ respectively. Two observations follow.
First, $A{=}\mathrm{random}$ ($24.32$ mIoU, $58.20\%$ ASR) is markedly
less destructive than ACI alone ($15.0$ mIoU, $66.76\%$ ASR), showing
that the CLIP-guided inversion in ACI extracts a structured semantic
dissimilarity signal that random destruction cannot reach. Second,
while $A{=}0$ ($3.49$) and random derangement ($4.14$) achieve mIoU
degradation comparable to ACI and SIP, they produce structurally
anomalous outputs: an all-zero affinity matrix would be flagged by
any consistency check, and a random label derangement violates the
semantic geometry of CLIP text embeddings. In contrast, ACI produces a
corrupted affinity matrix that preserves the row-stochastic structure
and sparsity statistics of the clean affinity, and SIP produces a
label map whose permutation is grounded in CLIP semantic geometry.
This comparison highlights that the contribution of CrACK is not raw
destructive power, which is partially achievable through trivial means
under our threat model, but a structured exploitation of the
inter-model interface that remains consistent with the pipeline's
expected output distribution.

\begin{figure}[t]
  \centering
  \includegraphics[width=1\linewidth]{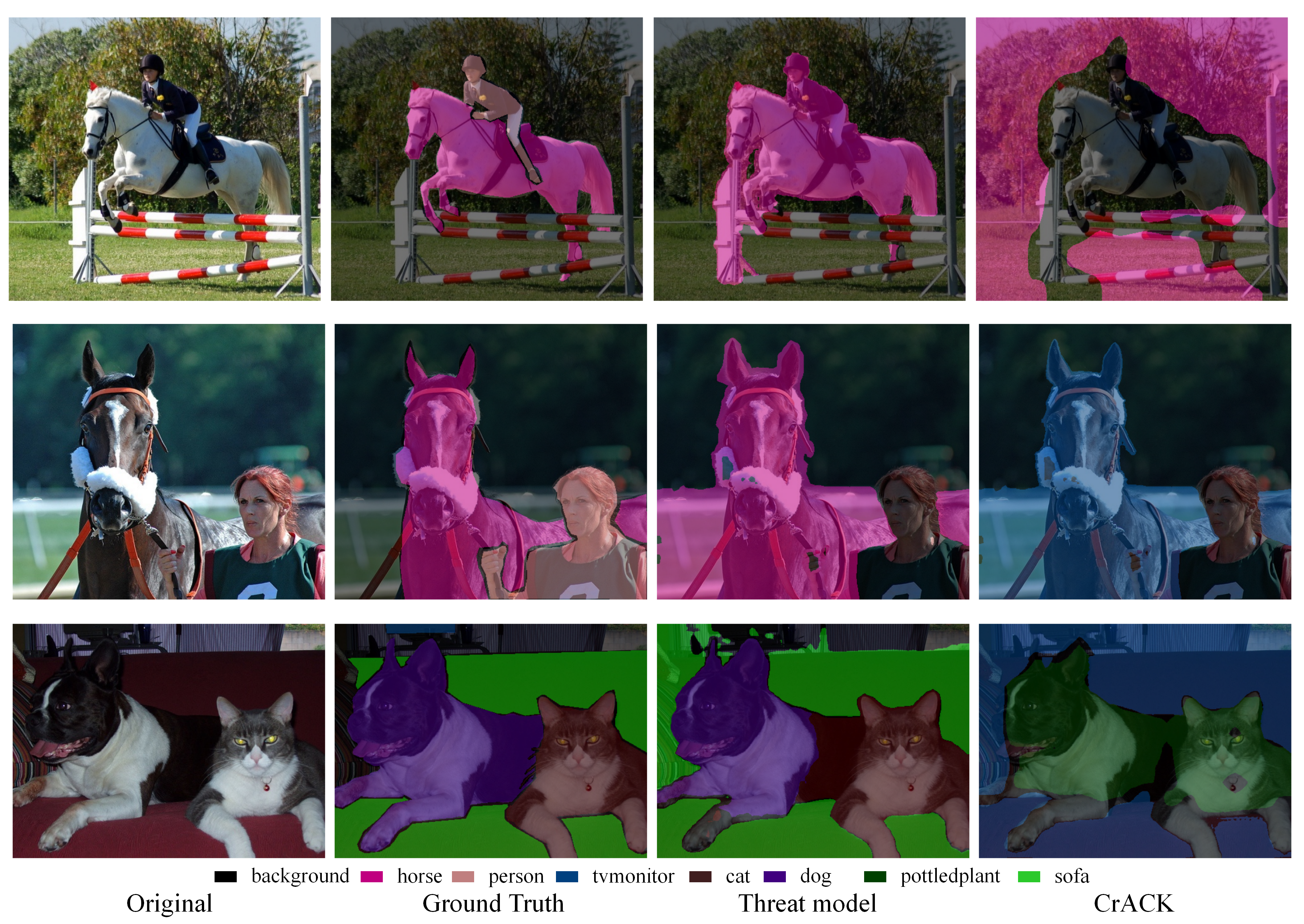}
  \vspace{1mm}
  \caption{Qualitative comparison of CrACK attack variants on Trident.
  From top to bottom we show segmentation outputs under ACI, SIP, and
  the full ACI plus SIP pipeline. The four columns show the input
  image, ground truth, clean prediction, and attacked prediction.}
  \label{fig:vis}
\end{figure}

Figure~\ref{fig:vis} provides qualitative results. ACI disrupts the
spatial coherence of the segmentation map, SIP redirects every
foreground label to the semantically most distant class, and the full
pipeline produces both effects simultaneously.

\begin{table}[t]
  \caption{Ablation of CrACK on Trident with PASCAL VOC21. The top
  block reports three trivial baselines that exploit the same threat
  model but use unstructured destruction. The bottom block reports the
  two attack stages of CrACK and their combination.}
  \vspace{1.5mm}
  \label{tab:ablation}
  \centering
  \small
  \begin{tabular}{lcc}
    \toprule
    Method & mIoU $\downarrow$ & ASR (\%) $\uparrow$ \\
    \midrule
    \multicolumn{3}{l}{\emph{Trivial glue-code baselines}}\\
    $A{=}0$                            & 3.49  & 99.60  \\
    $A{=}\mathrm{random}$              & 24.32 & 58.20  \\
    $Y{=}\mathrm{random\ derangement}$ & 4.14  & 100.00 \\
    \midrule
    \multicolumn{3}{l}{\emph{CrACK stages}}\\
    ACI only                           & 15.0 & 66.76  \\
    SIP only                           & 4.1  & 100.00 \\
    \rowcolor{ourshl}
    \textbf{ACI + SIP (Full CrACK)}    & \textbf{2.8} & \textbf{100.00} \\
    \bottomrule
  \end{tabular}
\end{table}

\section{Conclusion}
\label{sec:conclusion}
Training-free collaborative VFM pipelines are increasingly deployed
in safety-critical settings~\cite{koleilat2024medclipsam}, yet their
security has been assumed to follow from the robustness of their
individual models. We challenged this assumption by identifying the
semantic-spatial alignment dependency as a structurally universal
vulnerability of such pipelines, and proposed CrACK to exploit it at
inference time without touching any input pixel, model weight, or
training data. On four pipelines and eight benchmarks, CrACK reduces
Trident's average mIoU from $45.8$ to $1.55$ with ASR $98.2\%$, and
the corruption cascades into LLaVA-1.5-13B, dropping POPE from
$87.4$ to $58.3$, all while every VFM produces its exact clean
standalone output. These results indicate that per-model robustness
does not imply system robustness, and that the inter-model feature
interface must be treated as a first-class security boundary. Future
work includes black-box or transfer-based extensions of CrACK,
interface-level consistency defenses (e.g., real-time verification
between $C$ and $W$) that do not require retraining the underlying
VFMs, and extending the inter-model interface framing to multi-agent
LLM and tool-augmented systems~\cite{qi2024visual,gu2024agentsmith},
where intermediate outputs are exchanged in a structurally similar way.
 
\paragraph{Limitations.}
The current ACI formulation requires interception of SAM intermediate
features and read-only access to a CLIP patch-token extractor, matching
the supply-chain and microservice-API threat models in
Sec.~\ref{sec:threat} but not yet covering pure black-box settings.

\bibliographystyle{splncs04}
\bibliography{main}

\end{document}